\documentclass[conference]{IEEEtran}
\IEEEoverridecommandlockouts
\usepackage{graphicx}
\usepackage{amsmath}
\usepackage{algorithm}
\usepackage{algpseudocode}
\usepackage{cleveref}
\usepackage{booktabs}
\usepackage{adjustbox}
\usepackage{tabularx}
\usepackage{makecell}
\ifCLASSINFOpdf
\else
\fi
\begin{document}
%
\title{Class-balanced Open-set Semi-supervised Object Detection for Medical Images}


\author{
\IEEEauthorrefmark{1}
\IEEEauthorblockN{Zhanyun Lu}
\IEEEauthorblockA{Hangzhou Dianzi University\\
221050365@hdu.edu.cn}
\and
\IEEEauthorrefmark{1}
\IEEEauthorblockN{Renshu Gu}
\IEEEauthorblockA{Hangzhou Dianzi University\\
renshugu@hdu.edu.cn}
\and
\IEEEauthorblockN{Huimin Cheng}
\IEEEauthorblockA{Hangzhou Dianzi University\\
chm7777@hdu.edu.cn}
\and
\IEEEauthorblockN{Siyu Pang}
\IEEEauthorblockA{Hangzhou Dianzi University\\
21051706@hdu.edu.cn}
\and
\IEEEauthorblockN{Mingyu Xu}
\IEEEauthorblockA{Zhejiang University\\
xumingyu@zju.edu.cn}
\and
\IEEEauthorblockN{Peifang Xu}
\IEEEauthorblockA{Zhejiang University\\
xpf1900@zju.edu.cn}
\and
\IEEEauthorblockN{Yaqi Wang}
\IEEEauthorblockA{Communication University of Zhejiang\\
wangyaqi@cuz.edu.cn}
\and
\IEEEauthorblockN{Yuichiro Kinoshita}
\IEEEauthorblockA{University of Yamanashi\\
ykinoshita@yamanashi.ac.jp}
\and
\IEEEauthorblockN{Juan Ye}
\IEEEauthorblockA{Zhejiang University\\
yejuan@zju.edu.cn}
\and
\IEEEauthorrefmark{2}
\IEEEauthorblockN{Gangyong Jia}
\IEEEauthorblockA{Hangzhou Dianzi University\\
gangyong@hdu.edu.cn}
\and
\IEEEauthorrefmark{2}
\IEEEauthorblockN{Qing Wu}
\IEEEauthorblockA{Hangzhou Dianzi University\\
wuq@hdu.edu.cn}
}

\thanks{\IEEEauthorrefmark{1}These authors contributed equally to this work.}
\thanks{\IEEEauthorrefmark{2}Corredponding author.}
\maketitle
\begin{abstract}
Medical image datasets in the real world are often unlabeled and imbalanced, and Semi-Supervised Object Detection (SSOD) can utilize unlabeled data to improve an object detector. However, existing approaches predominantly assumed that the unlabeled data and test data do not contain out-of-distribution (OOD) classes. The few open-set semi-supervised object detection methods have two weaknesses: first, the class imbalance is not considered; second, the OOD instances are distinguished and simply discarded during pseudo-labeling. In this paper, we consider the open-set semi-supervised object detection problem which leverages unlabeled data that contain OOD classes to improve object detection for medical images. Our study incorporates two key innovations: Category Control Embed (CCE) and out-of-distribution Detection Fusion Classifier (OODFC). CCE is designed to tackle dataset imbalance by constructing a Foreground information Library, while OODFC tackles open-set challenges by integrating the ``unknown'' information into basic pseudo-labels. Our method outperforms the state-of-the-art SSOD performance, achieving a 4.25 mAP improvement on the public Parasite dataset. The code will be released upon acceptance.
\end{abstract}


%
\IEEEpeerreviewmaketitle

\section{Introduction}
%
%
%
%

Recent years have witnessed the rapid development of object detection research  \cite{ren2015faster,redmon2016you,liu2016ssd}. Fully supervised models require a substantial amount of annotated data, which is labor-intensive and time-consuming, especially in the realm of medical image analysis. Subsequently, semi-supervised object detection (SSOD) methods \cite{chen2021temporal,jeong2019consistency,li2022pseco,liu2021unbiased,sohn2020simple,xu2021end,zhou2021instant}, which could effectively leverage unlabelled data, have emerged and gained increasing attention. Related research has also found wide application in medical image analysis \cite{hermoza2022censor,li2021document,zhang2023self,albert2021using,lai2021hetero,li2021attent,yang2021weakly}. These methods generally assume that the data used for training and testing share the same categories. However, in the medical imaging domain, rare diseases can appear unexpectedly in the testing set without being present in the training set. Due to the diversity and complexity of medical images, even experienced doctors may misdiagnose some rare and difficult-to-distinguish conditions, which coincides with the open-set problem in computer vision.

For natural images, Liu et al. \cite{liu2022open} advocated using an offline DINO \cite{caron2021emerging} model as a filter to eliminate information of unknown categories in unlabeled images. Wang et al.\cite{wang2023online} proposes an online framework to distinguish and filter the out-of-distribution (OOD) instances from the in-distribution (ID) instances during pseudo-labeling. However, few works consider the open-set semi-supervised object detection (OSSOD) problem in medical imaging despite that the OSSOD problem is equally prominent and worth investigating in this area.

\begin{figure}[t]
    \centering
    \includegraphics[width=1\linewidth]{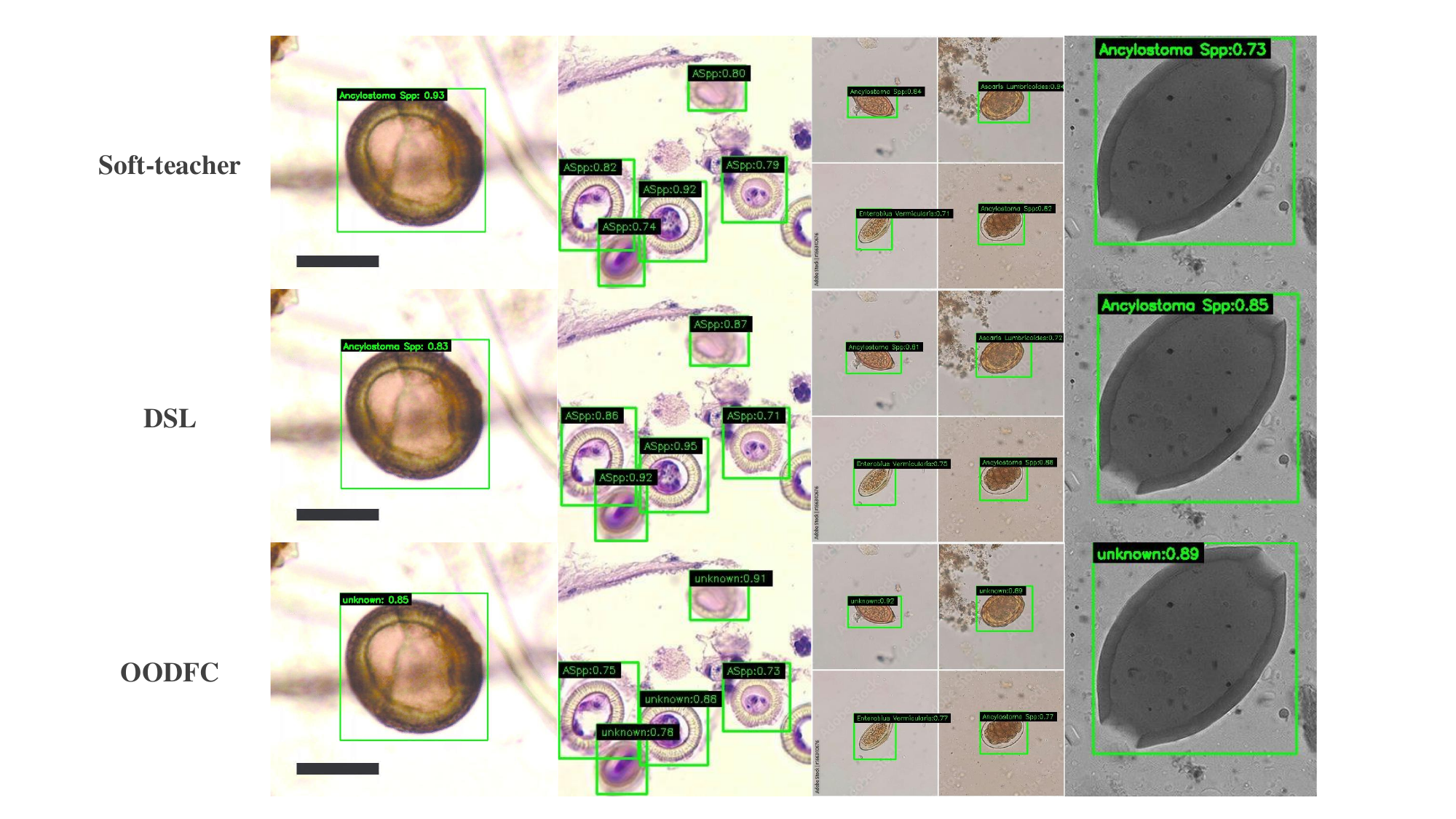}
    \caption{This is a visualization of the prediction results of three models on four images that contain unknown objects. In the predictions of soft-teacher and DSL, the unknown categories are incorrectly predicted as the Ancylostoma Spp category, which appears most frequently in the training set. In contrast, OODFC can produce ``unknown'' labels.}
    \label{fig:compare}
\end{figure}

In medical imaging, class imbalance is a very common issue, with a significant disparity in the number of common and rare classes. Unknown and rare categories are often misclassified by existing models as the most numerous known categories in the training set, meaning that models exhibit severe bias, resulting in high confidence in wrong predictions. Our research aims to fill this gap by proposing an innovative pipeline that makes the trained data more balanced. Another issue is that the performance of SSOD methods deteriorates when unknown categories are present. \cite{dhamija2020overlooked} points out that the existence of unknown categories will in turn seriously affect the model's ability to judge known categories. Previous OSSOD methods \cite{liu2022open} and \cite{wang2023online} distinguish and simply discard the unknown categories. In contrast, we propose to mitigate the interference of unknown categories on known categories by distinguishing and fusing the unknown class predictions.

In our framework, the Out-of-distribution Detection Fusion Classifier module (OODFC) incorporates the pseudo-labels generated by Opendet \cite{han2022expanding} which tags objects that it identifies as belonging to unknown categories with the label ``unknown'' as shown in Fig. \ref{fig:compare}. These pseudo-labels will be integrated into the basic pseudo-labels of SSOD framework as a hint to the model, aiding in the mitigation of the open-set problem.

On the other hand,  we designed the Category Control Embed (CCE) module targeted at mitigating the class imbalance problem. It dynamically constructs a Foreground Information Library to regulate the frequency of occurrence of various category objects in unlabelled data, thereby mitigating the class imbalance phenomenon.

Our contributions can be summarized as follows:
\begin{itemize}
 \item We propose the first open-set semi-supervised medical object detection framework, where a novel label fusion module is employed to reduce the interference of unknown categories on known ones. 
  \item Further, a novel category control embed module is integrated into the framework to alleviate the class imbalance issue in medical images. 
  \item Extensive experiments on private and public datasets show that the proposed method can significantly improve the performance in detecting known categories for medical images in an open-set semi-supervised setting.
 \end{itemize}

\begin{figure*}
    \centering
    \includegraphics[width=1\linewidth]{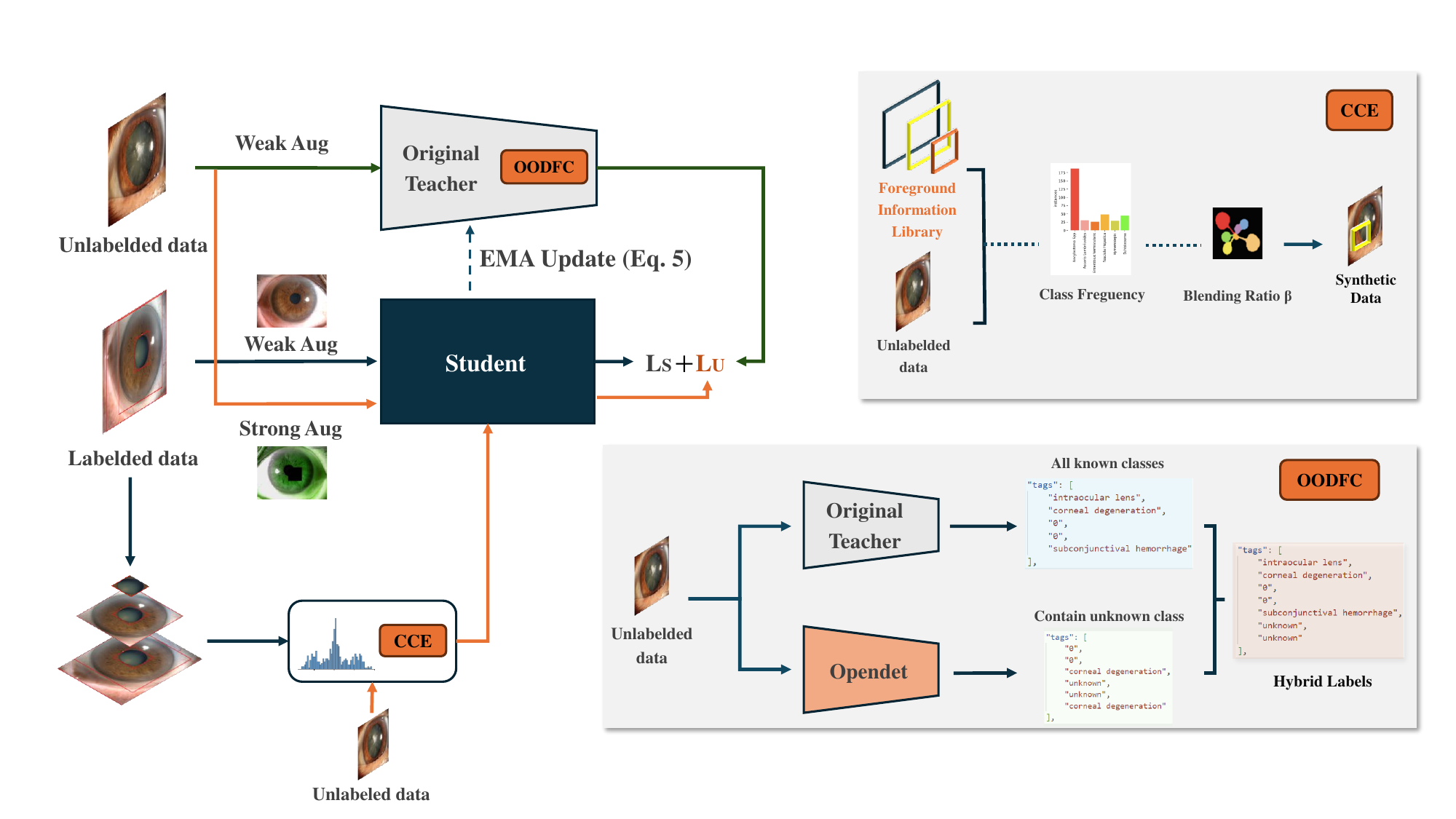}
    \caption{Overall structure of our framework. We designed two modules to address the class imbalance and open-set problems. The entire framework is based on a semi-supervised pipeline using a teacher-student network. Weak augmentation includes random flip, while strong augmentation includes random flip, color jittering, and cutout. The CCE mitigates category imbalance by dynamically embedding foreground information, while the OODFC prevents the model from misclassifying unknown categories as known ones by integrating unknown class information.}
    \label{fig:overall framework}
\end{figure*}

\section{Method}

As shown in Fig. \ref{fig:overall framework}, our framework includes the Category Control Embed (CCE) module to address the issue of class imbalance, and the Out-Of-Distribution Detection Fusion Classifier (OODFC) to integrate the unknown class information into basic pseudo-labels to mitigate interference from unknown categories. These two modules will be integrated into a basic Semi-Supervised Object Detection (SSOD) paradigm named Mean-Teacher \cite{liu2021unbiased,tarvainen2017mean,xu2021end,chen2022dense} pipeline. It will serve as the baseline, which consists of a semi-supervised network that utilizes ResNet50 \cite{he2016deep} as its backbone, FPN \cite{lin2017feature} as the neck, FCOS \cite{tian2019fcos} as the detector and EMA \cite{tarvainen2017mean} for updating the teacher network.

\subsection{Semi-supervised Object Detection}

Object detection is an essential task in computer vision that involves identifying and locating objects within images. Traditional methods for object detection require extensive annotated datasets, which are costly and time-intensive to generate. Semi-supervised object detection (SSOD) addresses this issue by utilizing both labeled and unlabeled data to enhance detection performance.

\subsubsection{Learning Framework}
Semi-supervised learning aims to leverage a large corpus of unlabeled data in conjunction with a smaller labeled dataset to boost model performance. Initially, the model is trained on the labeled data, and then it is used to predict labels for the unlabeled data. These predicted labels, or pseudo-labels, are subsequently used to fine-tune the model.

\subsubsection{Objective Function}

The objective function in SSOD typically combines a supervised loss on labeled data and an unsupervised loss on unlabeled data. The supervised loss \( L_s \) is calculated using standard object detection techniques like Faster R-CNN \cite{ren2015faster}, YOLO \cite{Jocher_Ultralytics_YOLO_2023}, or SSD \cite{liu2016ssd} on the labeled dataset \( D_l \). The unsupervised loss \( L_u \) leverages the pseudo-labels generated from the model's predictions on the unlabeled dataset \( D_u \).

Mathematically, the overall loss \( L \) can be expressed as:

\begin{equation}
L = L_s + \lambda L_u
\end{equation}

where \( \lambda \) is a weight factor that balances the contributions of the supervised and unsupervised losses. \( L_s \) is computed from the labeled data and includes components such as classification loss \( L_{cls} \) and localization loss \( L_{loc} \)

\begin{equation}
L_s = L_{cls} + L_{loc}
\end{equation}

Unsupervised Loss \( L_u \) is derived from the unlabeled data using pseudo-labels. It often includes consistency regularization and pseudo-labeling mechanisms. For instance, Mean-teacher \cite{tarvainen2017mean} or self-training methods can be employed to generate reliable pseudo-labels \(\hat{y}\) for the unlabeled data:

\begin{equation}
L_u = L_{consistency} + L_{pseudo}
\end{equation}

\subsubsection{Pseudo-Labeling}

Pseudo-labeling is a key component of SSOD. It firstly Train the model using the labeled data \( D_l \) and then applies the trained model to the unlabeled data \( D_u \) to predict pseudo-labels. It will filter the pseudo-labels based on confidence thresholds to ensure quality and use the filtered pseudo-labeled data to further train the model, updating its parameters iterative.

\subsubsection{Consistency Regularization}

Consistency regularization ensures that the model's predictions are stable and consistent under different perturbations (e.g., data augmentation). This technique encourages the model to produce similar predictions for the same image under various transformations.

\begin{equation}
L_{consistency}=\sum_{i=1}^N\|f_\theta(x_i)-f_\theta(T(x_i))\|^2
\end{equation}

where \( T(x_i) \) is a transformed version of \( x_i \), and \( f_\theta \) denotes the model's prediction function.

\subsection{EMA Mechanism}

In semi-supervised learning frameworks, we use Exponential Moving Average (EMA) mechanism \cite{tarvainen2017mean} to update model weights. This method stabilizes and improves the learning process of models, especially in scenarios with sparse or noisy labeled data.

The EMA \cite{tarvainen2017mean} update rule at time step \( t \) for model parameters \( \theta_{t} \) is defined as:

\begin{equation}
\theta_{t+1}=\alpha\theta_t+(1-\alpha)\theta_{t+1}'
\end{equation}

Here, \( \alpha \) is a decay coefficient between 0 and 1, controlling how much emphasis is placed on historical parameters. \(\theta_{t+1}^{\prime} \) represents the parameters of the model at the current time step.

In semi-supervised learning, there are typically a large number of unlabeled data points and a smaller set of labeled data points. EMA \cite{tarvainen2017mean} can smooth the update of model parameters by averaging over the entire dataset, particularly benefiting from the unlabeled data.

EMA \cite{tarvainen2017mean} reduces the variance in parameter updates, thereby stabilizing the model, which is crucial when training on a small amount of labeled data prone to noise. By smoothing parameter updates, it also improves the model's ability to generalize to unseen data, including unlabeled data. This is achieved by maintaining a stable model state throughout training. Furthermore, in semi-supervised settings, labeled data is often limited. EMA \cite{tarvainen2017mean} efficiently incorporates information from both labeled and unlabeled data to update model parameters, improving performance across the entire data distribution.

\subsection{OOD Detection Fusion Classifier}

OODFC is proposed to address the challenge of open-set semi-supervised object detection. 
Our framework introduces an additional ``unknown'' category label. This implies that when the detector encounters objects of unknown categories, it does not discard these data outright; instead, it labels them as ``unknown''. 

The advantage of this approach lies in its ability to preserve a greater amount of information, particularly for pseudo-labeled objects that may not have been correctly classified. Thus, the model can achieve a more comprehensive understanding and interpretation of data.

In the semi-supervised learning pipeline, a fully supervised teacher network \cite{tian2019fcos} is responsible for generating pseudo-labels for known categories, referred to as the original teacher. We use an additional detection module Opendet \cite{han2022expanding} to generate pseudo-labels for unknown categories. These two types of labels
are merged to form new hybrid labels as shown in Fig.~\ref{fig:overall framework} OODFC part. In the merging process, when two regions with IoU greater than 0.7 are predicted to be the known category $i$ and the unknown category respectively, we will use a novel dynamic threshold to filter ``unknown'' labels:

\begin{equation}
T_i = \max(0, \min(1, e^{\gamma \cdot (\text{AP}_i - 1)}))
\end{equation}

$T_i$ is the threshold used to determine whether to retain the label of the unknown category, which will be influenced by ${AP}_i$ while ${AP}_i$ represents the Average Precision for category $i$ of the fully supervised teacher network in the SSOD pipeline. $\gamma$ is a positive coefficient that controls the rate of growth of the function, which is set to 1.5. The choice of the exponential function is because the closer $AP_i$ is to 1, the thresholds need finer division. 

Although some areas may sometimes be marked as both a known and an unknown category, this method still plays a crucial role in guiding the model to recognize unknown objects, effectively reducing the interference of unknown categories on the known categories. The steps are shown as Algorithm \ref{alg:OODFC}

\begin{algorithm}
\caption{Pseudo code of OODFC}
\label{alg:OODFC}
\begin{algorithmic}[1]
\State \textbf{Input:} Labeled dataset $D_L$, Unlabeled dataset $D_U$, Threshold $T$
\State \textbf{Output:} Refined pseudo-labels for $D_U$
\State Train a fully supervised detector on $D_L$
\State Parallelly:
\State \quad Train original teacher network on $D_L$ and predict on $D_U$ to get pseudo-label1
\State \quad Train Opendet on $D_L$ and predict on $D_U$ to get pseudo-label2
\For{each image $x \in D_U$}
    \State Extract predictions for $x$ from pseudo-labels1 and pseudo-labels2
    \For{each object detection in pseudo-labels2}
        \If{category is 'unknown' \textbf{and} confidence $\geq T$}
            \State Append this detection as 'unknown' to the corresponding entry in pseudo-labels1
        \EndIf
    \EndFor
\EndFor
\State \textbf{return} Updated pseudo-labels1
\end{algorithmic}
\end{algorithm}

\begin{algorithm}[ht]
\caption{Pseudo code of CCE}
\label{alg:CCE}
\begin{algorithmic}[1]
\State \textbf{Input:} Labeled dataset $D_{labeled} = \{(x_i, y_i)\}_{i=1}^N$, Unlabeled dataset $D_{unlabeled}$
\State \textbf{Output:} Augmented dataset $D_{augmented}$

\Procedure{ForegroundInformationLibrary}{$D_{labeled}$}
    \State Compute class frequencies $\{f_c\}_{c=1}^C$ from $D_{labeled}$
    \State Extract bounding boxes $\{B_c\}_{c=1}^C$ from $D_{labeled}$
    \State Set $f_{target} = \frac{\sum_{c=1}^C f_c}{C}$
    \State Initialize augmented bounding boxes $\{B'_c\}_{c=1}^C \gets \emptyset$
    \For{each class $c$}
        \State Calculate $\alpha_c = \frac{f_{target}}{f_c}$
        \For{each bounding box $b \in B_c$}
            \State Augment $b$ to $\{b'_j\}$ based on $\alpha_c$
            \State Add $\{b'_j\}$ to $B'_c$
        \EndFor
    \EndFor
    \State \textbf{return} $\{B'_c\}_{c=1}^C$
\EndProcedure

\Procedure{BalanceAndAugment}{$D_{labeled}$, $D_{unlabeled}$}
    \State Initialize $D_{augmented} \gets \emptyset$
    \State $\{B'_c\}_{c=1}^C \gets$ \Call{ForegroundInformationLibrary}{$D_{labeled}$}
    \For{each class $c$}
        \For{each augmented bounding box $b'_j \in B'_c$}
            \State Blend $b'_j$ with a random image $u$ from $D_{unlabeled}$ using $\beta$ ratio
            \State Add blended image to $D_{augmented}$
        \EndFor
    \EndFor
    \State \textbf{return} $D_{augmented}$
\EndProcedure
\end{algorithmic}
\end{algorithm}

\subsection{Category Control Embed}
Category Control Embed (CCE) is the second module we proposed. To address the common issue of class imbalance in object detection tasks, we propose a solution based on image fusion, inspired by the classic Mixup \cite{zhang2017mixup} method. In our consideration, class imbalance refers not only to the disparity in the number of images but more specifically to the imbalance in the number of foreground object categories within the images. To tackle this challenge, we create a Foreground Information Library. This library contains segments of each bounding box from labeled images along with their corresponding annotation information. In a dataset with class imbalance, there is a significant difference in the number of segments from different classes in the library. This library will then selectively dynamically enhance the segments to adjust the quantity of foreground information from various categories, aiming for a more balanced representation. The steps are shown in Algorithm \ref{alg:CCE}

In detail, To balance the foreground information library, we use oversampling for categories below the average quantity and undersampling for those above (Algorithm \ref{alg:CCE}, Line 6).

This balanced Foreground Information Library will then be integrated into unlabeled data, generating a new set of synthetic datasets. During the addition process, a parameter $\beta$ is used to control the fusion ratio between two images, achieving varying degrees of fusion effects as well as avoiding the loss of valuable information on the overlaid medical image: 

\begin{equation}
\begin{aligned}x_i^{Syn}&=\beta b_j^{'}+(1-\beta)x_i,\\y_i^{Syn}&=combine(y_i,y_j)=y_j,\end{aligned}
\end{equation}

Where $b_j^{'}$ is an augmented foreground segment. $x_i$ is the part that is randomly selected in an unlabeled image, and this part will be fused with $b_j^{'}$; $y_i$ is the empty annotation of unlabeled data, so the $y_i^{Syn}$ finally becomes $y_j$, which is the label of $b_j^{'}$. $\beta$ is set to 0.5, which means the images are blended with equal weights. 

\begin{figure*}
    \centering
    \includegraphics[width=0.9\linewidth]{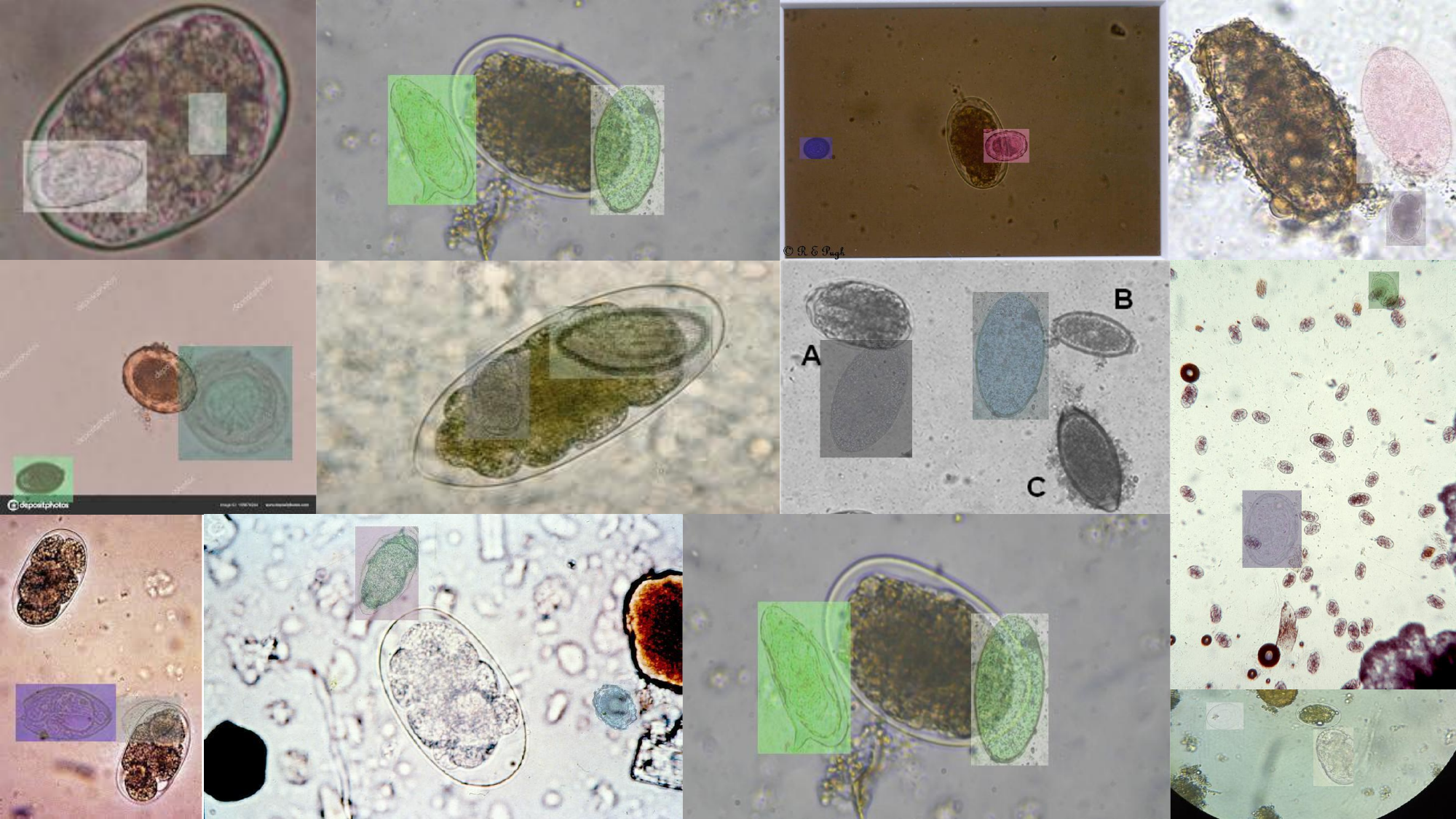}
    \caption{\textbf{Visualization of synthetic images generated by CCE.} In these examples, several foreground image segments are randomly added to the original unlabeled image.}
    \label{fig:cce example}
\end{figure*}

The foreground parts' position is a random coordinate within the unlabeled image boundaries. The synthetic data label includes the foreground category and coordinates, with bounding box confidence set to 0.8. Some examples are shown in Fig.~\ref{fig:cce example}. An augmentation factor $\alpha_c$ is used for the extent of augmentation required for each class to achieve balance. By selecting bounding box regions, this method specifically alleviates the imbalance in the number of foreground objects across classes. This entire procedure is named CCE, which will be integrated into Mean-Teacher \cite{tarvainen2017mean} and executed automatically during runtime.

\section{Experiments and results}

\subsection{Datasets}

In this study, we applied the framework to assess its performance on two different datasets: a private Ophthalmology dataset and a public Parasite dataset \cite{parasites-1s07h_dataset}. Both datasets inherently exhibit class imbalance, characterized by significant differences in the number of images across different categories. The two datasets used in our study contain quite distinct contents, verifying the model's ability to generalize across different types of medical imaging data.

\noindent \textbf{The Slit Lamp Ophthalmology Dataset.} The private Slit Lamp Ophthalmology Dataset includes 11 known categories and three unknown categories: ``Conjunctival Congestion'', ``Limb Conjunctival Tumor'', and ``Keratitis''. These three categories are not involved in the training process at all and only appear in the validation set. From the annotated data of the 11 known categories, we selected 10\% from each category as the labeled training data. This data, along with the unlabeled training data, forms a semi-supervised training set. The remaining 90\% of the labeled data and all the data of the three unknown categories are used as the test set. 

This private dataset comprises 136 labeled training images, 1224 unlabeled training images, and 1782 validation images, adopting a classic 10\% labeled data and 90\% unlabeled data split for the training set. The unlabeled data comes from an external validation set from a hospital. The external validation set and the internal training set come from different hospitals, and the lighting conditions may vary. This poses a greater challenge to the model's generalizability.

\noindent \textbf{The Parasite dataset.} Similarly, in the public Parasite dataset, we have 6 known categories plus two unknown parasite categories: ``Taenia Sp'' and ``Trichuris Trichiura''. These two unknown categories also do not appear in the training process and are only used for performance evaluation. From the 6 known categories, we also took 10\% from each as the labeled training data, which, along with the unlabeled data, constitutes the semi-supervised training set. The original validation set of this dataset, along with all the data of the two unknown categories, is used as the test set. 

The public Parasite dataset includes 152 labeled training images, 1332 unlabeled training images, and 411 validation images. The unlabeled and labeled data also comprise 10\% and 90\% in the training set, respectively. The detailed information about Categories and Splits is shown in Table \ref{tab:c and s}

\begin{table}[h]
    \centering
    \caption{Categories and Splits of the Datasets}
    \label{tab:c and s}
    \begin{tabularx}{\linewidth}{>{\raggedright\arraybackslash}X|>{\raggedright\arraybackslash}p{0.15\linewidth}|>{\raggedright\arraybackslash}p{0.15\linewidth}|>{\raggedright\arraybackslash}p{0.15\linewidth}}
        \Xhline{3\arrayrulewidth}
        \textbf{Dataset} & \textbf{Train} & \textbf{Val} & \textbf{Total} \\ 
        \hline
        \multicolumn{4}{l}{\textbf{Private Ophthalmology Dataset}} \\
        \hline
        \textbf{Known Category Foreground Objects Number} \\
        \hline
        0 & 375 & 4829 & 5204 \\ 
        Corneal Degeneration & 28 & 266 & 294 \\ 
        Pigmented Nevus & 46 & 411 & 457 \\ 
        Pterygium & 8 & 82 & 90 \\ 
        Cataract & 10 & 96 & 106 \\ 
        Conjunctival Cyst & 10 & 96 & 106 \\ 
        Subconjunctival-Hemorrhage & 19 & 183 & 202 \\ 
        Artificial Lens & 6 & 62 & 68 \\ 
        Palpebral Fissure Spots & 21 & 188 & 209 \\ 
        Corneal Scar & 5 & 53 & 58 \\ 
        Lens Dislocation & 21 & 27 & 48 \\ 
        \hline
        \textbf{Unknown Categories} \\
        \hline
        Conjunctival Congestion & - & - & - \\ 
        Limb Conjunctival Tumor & - & - & - \\ 
        Keratitis & - & - & - \\ 
        \hline
        \textbf{Image number} \\
        \hline
        Labeled Training Images & 136 & 0 & 136 \\ 
        Unlabeled Training Images & 1224 & 0 & 1224 \\ 
        Validation Images & 0 & 1782 & 1782 \\ 
        \Xhline{3\arrayrulewidth}
        \multicolumn{4}{l}{\textbf{Public Parasite Dataset}} \\
        \hline
        \textbf{Known Category Foreground Objects Number} \\
        \hline
        Ancylostoma Spp & 187 & 140 & 327 \\ 
        Ascaris Lumbricoides & 30 & 115 & 145 \\ 
        Enterobius Vermicularis & 26 & 154 & 180 \\ 
        Fasciola Hepatica & 48 & 76 & 124 \\ 
        Hymenolepis & 28 & 89 & 117 \\ 
        Schistosoma & 45 & 90 & 135 \\ 
        \hline
        \textbf{Unknown Categories} \\
        \hline
        Taenia Sp & - & - & - \\ 
        Trichuris Trichiura & - & - & - \\ 
        \hline
        \textbf{Image number} \\
        \hline
        Labeled Training Images & 152 & 0 & 152 \\ 
        Unlabeled Training Images & 1332 & 0 & 1332 \\ 
        Validation Images & 0 & 411 & 411 \\ 
        \Xhline{3\arrayrulewidth}
    \end{tabularx}
\end{table}

\subsection{Metrics}
We report the Average Precision (AP) for each class and mean Average Precision (mAP) for all classes, following object detection conventions \cite{Jocher_Ultralytics_YOLO_2023}. AP (Average Precision) measures the accuracy of a model's predictions in object detection by considering both precision and recall. It is calculated as the area under the precision-recall curve. mAP (Mean Average Precision) is the average of AP scores across all classes in a dataset, providing a single metric for evaluating overall model performance across multiple classes. In our experiment, we use AP50 as our evaluation metric, which means we measure the Average Precision with an Intersection over Union (IoU) threshold of 50\%. This indicates that a predicted bounding box is considered a true positive if it overlaps with the ground truth bounding box by at least 50\%.

\subsection{Implementation Details}

In our experiment, the baseline semi-supervised pipeline is Mean-Teacher \cite{tarvainen2017mean}. It's worth noting that both the Opendet \cite{han2022expanding} detector and the Mean-Teacher \cite{tarvainen2017mean} pipeline use the same batch of labeled training data. Therefore, no additional labeled data was utilized. 

The model's training parameters are lr=0.001, momentum=0.9, weight-decay=0.0001, warmup-iters=4000. Weak augmentation includes random flip, while strong augmentation includes random flip, color jittering, and cutout. For other state-of-the-art (SOTA) models used for comparison we employ their default parameters in the publicly available code. 

\begin{table*}[ht]
\centering
\caption{Accuracy on private ophthalmology dataset: 1. Artificial Lens, 2. Cataract, 3. Corneal Degeneration, 4. Corneal Scar, 5. 0, 6. Conjunctival Cyst, 7. Lens Dislocation, 8. Palpebral fissure spots, 9. Pterygium, 10. Subconjunctival Hemorrhage, 11. Pigmented Nevus. We indicated the highest AP value with bold, and the second highest AP value with an underline. }
\label{tab:per AP of private}
\begin{adjustbox}{width=0.9\textwidth}
\begin{tabular}{@{}lcccccccccccc@{}}
\toprule
Model/Category & 1 & 2 & 3 & 4 & 5 & 6 & 7 & 8 & 9 & 10 & 11 & mAP \\ 
\midrule
Percentage(\%) & 1.13 & 1.89 & 5.28 & 0.94 & 70.75 & 1.89 & 0.38 & 3.96 & 1.51 & 3.58 & 8.68 & - \\
\midrule
FCOS \cite{tian2019fcos} & \underline{62.1} & \textbf{82.7} & 67.9 & 19.1 & 69.2 & 22.7 & 35.4 & 33.7 & 53.0 & 18.5 & 42.6 & 46.08 \\
Opendet \cite{han2022expanding} & - & - & - & - & - & - & - & - & - & - & - & 47.51 \\
YOLOv8 \cite{Jocher_Ultralytics_YOLO_2023} & \textbf{66.4} & \underline{81.0} & \underline{81.4} & 14.7 & \textbf{92.8} & 22.1 & \textbf{42.4} & 31.6 & 58.6 & \underline{30.4} & 55.2 & \underline{52.42} \\
\midrule
Soft-teacher \cite{xu2021end} & 39.9 & 72.6 & \textbf{84.7} & \underline{18.7} & 70.3 & \underline{31.3} & 40.6 & 30.5 & 60.1 & \textbf{36.6} & 53.9 & 49.02 \\
DSL \cite{chen2022dense} & 38.9 & 78.4 & 68.5 & 17.6 & 85.5 & 29.4 & \underline{41.3} & \underline{48.8} & \underline{64.4} & 22.7 & \underline{58.0} & 50.31 \\
ConsistentTeacher \cite{wang2023consistent} & 50.2 & 79.0 & 69.5 & 16.3 & 85.2 & 23.6 & 39.7 & 37.4 & 52.0 & \underline{30.4} & 48.4 & 48.34 \\
\midrule
Ours & 48.7 & 78.3 & 72.4 & \textbf{20.9} & \underline{85.8} & \textbf{32.1} & 41.1 & \textbf{52.2} & \textbf{67.0} & 24.2 & \textbf{62.6} & \textbf{53.21} \\
\bottomrule
\end{tabular}
\end{adjustbox}
\end{table*}

\begin{table*}[ht]
\centering
\caption{Accuracy on the public Parasite dataset: 1. Ancylostoma Spp, 2. Ascaris Lumbricoides, 3. Enterobius Vermicularis, 4. Fasciola Hepatica, 5. Hymenolepis, 6. Schistosoma Cyst.}
\label{tab:per AP of public}
\begin{adjustbox}{width=0.7\textwidth}
\begin{tabular}{@{}lccccccccc@{}}
\toprule
Model/Category & 1 & 2 & 3 & 4 & 5 & 6 & mAP\\ 
\midrule
Percentage(\%) & 51.37 & 8.24 & 7.14 & 13.19 & 12.36 & 7.69 & -\\
\midrule
FCOS \cite{tian2019fcos} & 54.6 & 58.1 & 40.6 & \underline{37.2} & 45.9 & 47.2 & 47.27\\
Opendet \cite{han2022expanding} & - & - & - & - & - & - & 48.13\\
YOLOv8 \cite{Jocher_Ultralytics_YOLO_2023} & \textbf{74.7} & 58.7 & 43.7 & 34.3 & 44.1 & 51.6 & 51.20\\
\midrule
Soft-teacher \cite{xu2021end} & \underline{73.3} & \underline{66.8} & \underline{44.8} & \underline{39.1} & 58.6 & 33.4 & 52.67\\
DSL \cite{chen2022dense} & 54.1 & 52.9 & 43.1 & 32.3 & \textbf{68.8} & \underline{64.3} & 52.58\\
Consistent-Teacher \cite{wang2023consistent} & 55.4 & \textbf{72.7} & 20.2 & \textbf{51.5} & \underline{63.5} & 55.3 & \underline{53.10}\\
\midrule
Ours & 65.1 & 63.7 & \textbf{50.5} & 30.5 & 61.9 & \textbf{72.4} & \textbf{57.35}\\
\bottomrule
\end{tabular}
\end{adjustbox}
\end{table*}

\subsection{Results}

In our research, we adopted the following methods to ensure the accuracy and reliability of the results: In a single experiment, we recorded the highest mean Average Precision (mAP) and repeated the experiment five times to eliminate random errors. The experimental results are the average of these five experiments. To make effective comparisons, We compared our method with the state-of-the-art fully supervised object detection models and semi-supervised object detection models. It should be noted that the previous work by Liu et al. \cite{liu2022open} has not been open-sourced. Hence, we were unable to make a comparison. 

In Table.~\ref{tab:per AP of private}, we presented the accuracy of different models on each class and mAP of all classes in the private dataset (Note: `0' means no disease). Meanwhile, Table.~\ref{tab:per AP of public} shows the results of the same experimental setup using the public Parasite datasets. Percentage means the number of objects in this category as a percentage of the total and is used to show whether the current category belongs to the majority category or the minority category. The models in the upper half of the table are fully supervised models, and the lower half are semi-supervised models. In these tables, we indicated the highest AP value with bold, and the second highest AP value with an underline.

\subsection{Discussion}

In our experiments, the proposed model achieved the highest mean Average Precision (mAP) on both datasets. This performance surpasses the state-of-the-art semi-supervised models and high-performance fully supervised models with various architectures. This achievement demonstrates the effectiveness of our approach in leveraging the unique advantages of the proposed modules.

A detailed examination of the AP for each category reveals that our model excels particularly in minority categories. While traditional methods tend to overlook these categories due to the inherent class imbalance problem in many datasets, our approach addresses this issue to a certain extent. By incorporating CCE that specifically enhances the number of minority classes, our model ensures that these categories receive adequate attention during the training process.

Additionally, although recent large models (such as SAM \cite{kirillov2023segment}, ViT \cite{dosovitskiy2020image} and CLIP \cite{radford2021learning}) exhibit exceptional performance across different tasks, our model has several advantages compared to these powerful models:
\begin{itemize}
    \item \textbf{Data efficiency}: Specialized models adapt better to limited or lower-quality data.
    \item \textbf{Real-time requirements}: Lightweight models excel in scenarios requiring instant decisions due to lower computational overhead.
    \item \textbf{Compatibility}: This component integrates with any pseudo-label-based semi-supervised object detection model to reduce interference from unknown classes.
\end{itemize}

\subsection{Ablation study}
As mentioned, we adopted a basic Semi-Supervised Object Detection (SSOD) paradigm as our baseline, named Mean-Teacher \cite{tarvainen2017mean}. This serves as the baseline for comparison against the model that incorporates the CCE and OODFC modules respectively and that incorporates both two modules. Ablation study on two datasets is shown in Table \ref{tab:ablation_combined}. Experimental results show the model with both CCE and OODFC consistently achieves the highest accuracy, thereby validating their effectiveness.

\begin{table}[h]
    \centering
    \caption{Ablation study on two datasets}
    \begin{tabular}{ccc}
    \toprule
    Methods & mAP\(_{50}\) (Ophthalmology) & mAP\(_{50}\) (Parasite) \\
    \midrule
    \multicolumn{1}{l}{Mean-Teacher \cite{tarvainen2017mean}} & 48.38 & 52.05 \\
    \multicolumn{1}{l}{+CCE} & 50.92 & \underline{55.50} \\
    \multicolumn{1}{l}{+OODFC} & \underline{52.44} & 53.79 \\
    \multicolumn{1}{l}{+CCE +OODFC (Ours)} & \textbf{53.21} & \textbf{57.35} \\
    \bottomrule
    \end{tabular}
    \label{tab:ablation_combined}
\end{table}


\section{Conclusions and Future Work}
This study presents a pioneering approach in semi-supervised object detection for medical imaging, tackling the open-set problem and the class imbalance challenge. We introduce two novel modules:  OODFC, which integrates unknown class label information to minimize unknown classes' interference on known category judgments; and CCE, which dynamically embeds foreground information to reduce bias towards majority classes. Our model outperforms state-of-the-art semi-supervised as well as fully-supervised object detection methods on two datasets (a public Parasite dataset and a private ophthalmology dataset), demonstrating excellent generalization ability.

We used Opendet \cite{han2022expanding} as the ``unknown'' class information producer, yet our framework design is adaptable to include any model capable of detecting unknown categories. This allows for flexible replacement with various external detectors to enhance the model's ability to recognize unknown categories, whether the detector is fully supervised or semi-supervised. Future research could evaluate the performance of different external detectors in our framework.

\section{Acknowledgments}
This research was supported by the National Natural Science Foundation Regional Innovation and Development Joint Fund (No. U20A20386), the National Natural Science Foundation of China (No. U22A2033, No. 62202130 and No. 62206242), National Key R\&D Projects (No. 2022YFE0210700), the Zhejiang Provincial Science and Technology Program in China (No. LQ22F020026).

\bibliographystyle{IEEEtran}







%




\end{document}